\documentclass[letterpaper, 10 pt, conference]{ieeeconf}  

\IEEEoverridecommandlockouts                              

\usepackage{graphicx} 
\usepackage{tensor}
\usepackage[font=footnotesize]{caption}
\usepackage{cite}
\usepackage[font=normal]{subcaption}
\usepackage{booktabs}
\usepackage{times} 
\usepackage{bm}
\usepackage{amsmath,amssymb,mathrsfs,mathdesign,upgreek,dsfont,gensymb}
\usepackage[ruled,vlined]{algorithm2e}
\usepackage{verbatim}
\usepackage{diagbox}
\usepackage{cleveref}
\usepackage{todonotes}
\usepackage{soul}
\usepackage[multiple]{footmisc}
\usepackage{multirow}
\usepackage{threeparttable}

\graphicspath{{./figures/}}

\DeclareCaptionSubType*[Alph]{table}
\DeclareCaptionLabelFormat{mystyle}{Table~\bothIfFirst{#1}{ ̃}#2}
\def\code#1{\texttt{#1}}
\DeclareMathOperator*{\argmin}{arg\!\min}

\newcommand*\diff{\mathop{}\!\mathrm{d}}

\newcommand \ffsomp {FFS-OMP }
\newcommand \figref{Figure~\ref}
\newcommand \tabref{Table~\ref}
\newcommand \secref{Section~\ref}
\newcommand \algref{Algorithm~\ref}

\title{\LARGE \bf
Optimal Motion Planning using Finite Fourier Series \\ in a Learning-based Collision Field
}

\author{Feng Yichang, Wang Jin$^{*}$, Lu Guodong 
\thanks{$^{*}$ Wang Jin, corresponding author,  1) State Key Laboratory of Fluid Power and Mechatronic Systems, School of Mechanical Engineering,
Zhejiang University, Hangzhou 310027, China;  2) Engineering Research Center for Design Engineering and Digital Twin of Zhejiang Province, School of Mechanical Engineering,
Zhejiang University, Hangzhou 310027, China. 
        {\tt\small dwjcom@zju.edu.cn}}%
}

\begin{document}

\maketitle
\thispagestyle{empty}
\pagestyle{empty}

\begin{abstract}

This paper utilizes finite Fourier series to represent a time-continuous motion and proposes a novel planning method that adjusts the motion harmonics of each manipulator joint. Primarily, we sum the potential energy for collision detection and the kinetic energy up to calculate the Hamiltonian of the manipulator motion harmonics. Though the adaptive interior-point method is designed to modify the harmonics in its finite frequency domain, we still encounter the local minima due to the non-convexity of the collision field. In this way, we learn the collision field through a support vector machine with a Gaussian kernel, which is highly convex. The learning-based collision field is applied for Hamiltonian, and the experiment results show our method's high reliability and efficiency. 

\end{abstract}

\section{INTRODUCTION}

\begin{figure}[htb]
	\begin{centering}
		\begin{subfigure}[t]{0.265\textwidth}
			\centering
			\includegraphics[width=1\linewidth]{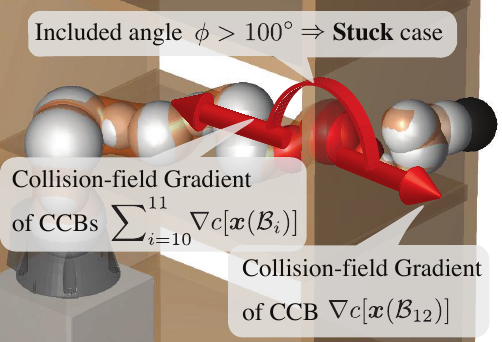}
			\caption{Stuck case in space scale}
			\label{fig:stuck-space}
		\end{subfigure}
		\hfill
		\begin{subfigure}[t]{0.205\textwidth}
			\centering
			\includegraphics[width=1\linewidth]{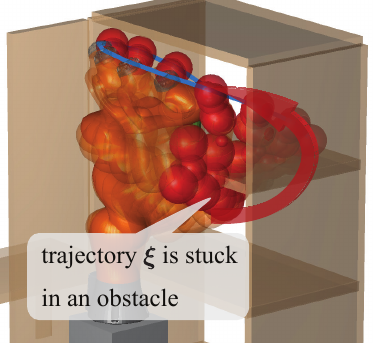}
			\caption{Stuck case in time scale}
			\label{fig:stuck-time}
		\end{subfigure}
	\end{centering}
\protect\caption{
The grey CCBs assemble an LBR-iiwa, while the red denotes an in-collision case. The stuck case occurs when the included angle $\phi > \phi_\text{tol}$.
\label{fig:stuckCase}}
\end{figure}

Manipulators play an indispensable role in industrial intelligence and take on the missions of sorting, grasping, carrying, etc., in some complex scenarios scattered intensely by obstacles. Therefore, a highly reliable and efficient motion planner is urged for safe execution. Though some numerical methods \cite{Zucker2013CHOMP, Mukadam2018GPMP, Bhardwaj2020dGPMP, Schulman2013SCO} search for the solution informed by gradients to improve the solving efficiency of the sampling-based optimization methods \cite{Kavraki1998PRMs, Karaman2011RRT*-PRM*, Rajendran2019CODES}, they still easily encounter the local minima due to the high convexity of the problem. Therefore, some sampling methods \cite{mandalika2019GLS, ichter2018learning-SBMP} modify the searching tree with the cut branches and restricted searching space to overcome this problem faster, regardless of their relatively high Hamiltonian. Our former study \cite{feng2022iSAGO} proposes the definition of the stuck cases in time and space scales, as shown in \figref{fig:stuckCase}, both of which lead to the local minima or the unsafe failure planning and introduce stochastic gradient descent to improve the success rate with high efficiency. However, its reliability is still weakened when the obstacles wrap the feasible solution up. 

This paper utilizes finite Fourier series (FFS) to represent the time-continuous motion and adjust the harmonics to make an optimal motion planning (\ffsomp): 

(i) To address the stuck case in the time scale, we introduce FFS, composed of finite harmonics instead of a discrete waypoint series, to represent the manipulator motion. According to our former studies \cite{feng2022iSAGO, yichang2021AGP-STO}, we have found that some trajectories with low Hamiltonian cost actually collide with the obstacles because some adjacent waypoints are separated so far that even the mid-waypoints are collision-free. In this way, FFS can narrow this separation down through a small number of harmonics since the large separation, meaning the straight line motion, needs an infinite number of harmonics to represent. 

(ii) As for the computational efficiency, we first linearize the Hamiltonian and use its functional gradient to approximate the Hessian matrix similar to the Gauss-Newton method. Considering the complex interaction between the manipulator and its surrounding obstacles, we adopt the adaptive method \cite{Kingma2014Adam} to modify the linearized Hamiltonian in account of the history gradient information of the potential functional used for collision check. Then, the adaptive interior point method is applied to gain the gradient descent under constraints, such as joint limits and start and goal conditions.

(iii) To solve the stuck case in the space scale, the support vector machine (SVM) \cite{Boser1992SVM} is introduced to find the safety boundary described by the support vectors. Since the Gaussian kernel is convex and the boundary excludes the deep stuck case as shown in \figref{fig:stuckCase}, the learning-based collision field is convex in the 3D Euclidean space, which could somewhat help to release the non-convexity of that in the configuration space, which is validated by the experiment results with high success rate. Unlike \cite{Das2020iSVM} learning in the configuration space, our learning in the Euclidean space requires a relatively more minor training set to gain sophisticated information so that the total learning period is only a few seconds.  

The experiment of the 15 planning tasks with 44 problems on LBR-iiwa and AUBO-i5 (Section~\ref{sec:Evaluation}) validates higher reliability of \ffsomp than the numerical methods~\cite{Zucker2013CHOMP, Mukadam2018GPMP, Schulman2013SCO} and its higher efficiency than the sampling methods~\cite{Kalakrishnan2011STOMP, Kuffner2000RRT-connect}.

\section{RELATED WORKS}

\subsection{Optimization based Motion Planning}
The main concern of numerical optimization is rapidly descending to an optimum.  CHOMP~\cite{Zucker2013CHOMP}, ITOMP~\cite{Park2012ITOMP}, and GPMP~\cite{Mukadam2018GPMP} adopt the gradient descent method with the fixed step size. CHOMP uses Hamiltonian Monte Carlo (HMC)~\cite{Shirley2011HMC} for success rate improvement. To lower the computational cost, GPMP and dGPMP~\cite{Bhardwaj2020dGPMP} adopt iSAM2~\cite{Kaess2012iSAM2} to do incremental planning, and each sub-planning converges with super-linear rate via Levenberg–Marquardt (LM)~\cite{Levenberg1944LM} algorithm. Meanwhile, ITOMP interleaves planning with task execution in a short-time period to adapt to the dynamic environment. Moreover, TrajOpt~\cite{Schulman2013SCO} uses the trust-region~\cite{Byrd2000TrustRegion} method to improve efficiency. It is also adopted by GOMP~\cite{Ichnowski2020GOMP} for grasp-optimized motion planning with multiple warm restarts learned from a deep neural network. Instead of deep learning, ISIMP~\cite{Kuntz2020ISIMP} interleaves sampling and interior-point optimization for planning. Nevertheless, the above methods may converge to local minima when the objective function is not strongly convex. 

\subsection{Sampling based Motion Planning}
Unlike the numerical method, the sampling method constructs a search graph to query feasible solutions or iteratively samples trajectories for motion planning. 

PRM~\cite{Kavraki1998PRMs} and its asymptotically-optimal variants like PRM*~\cite{Karaman2011RRT*-PRM*} and RGGs~\cite{Solovey2018RGGs} make a collision-free connection among the feasible vertexes to construct a roadmap. Then they construct an optimal trajectory via shortest path (SP) algorithms like Dijkstra~\cite{Dijkstra1959Dijkstra-Alg} and Chehov~\cite{Hofmann2015Chehov}, which store and query partial trajectories efficiently. Unlike PRM associated with SP, RRT~\cite{LaValle1998RRTs} and its asymptotically-optimal variants like RRT*~\cite{Karaman2011RRT*-PRM*} and CODEs~\cite{Rajendran2019CODES} find a feasible solution by growing rapidly-exploring random trees (RRTs). 

STOMP~\cite{Kalakrishnan2011STOMP} resamples trajectory obeying Gaussian distribution renewed by the important samples. \cite{Petrovic2019HGP-STO} introduces the GPMP's function to improve the searching efficiency of STOMP. Moreover, SMTO~\cite{Osa2020SMTO} applies Monte-Carlo optimization for multi-solution and refines them numerically. 

\subsection{Machine Learning method}

Recent studies have introduced machine learning methods to learn prior knowledge in motion planning. 

iSVM \cite{Das2020iSVM} learns a feasible configuration space incrementally based on the support vector machine (SVM) to elevate the collision-check efficiency. Simultaneously, \cite{Salehian2018MultiArmCoord} introduces SVM for multi-arm collision-check during coordinated motion planning. 
\cite{Lembono2020WS-TO} builds a motion memory and utilizes it for the warm start of the numerical methods by the Gaussian process regression (GPR), while dGPMP \cite{Bhardwaj2020dGPMP} learns the factor graph covariances for the Gaussian process motion planning. However, their generalization ability is limited by the simplicity and linearity of GPR. So \cite{Osa2022NN-TO, Shi2022NN-WS-TO} implement the neural network for a warm start to improve the planner reliability in various scenarios. 
Except for the implementation in the numerical method, \cite{ichter2018learning-SBMP} learns the sampling distribution to improve the planning efficiency of the asymptotical sampling methods, such as RRT* and PRM*.

\section{FINITE FOURIER SERIES FOR TRAJECTORY REPRESENTATION}\label{sec:ffs}

\begin{figure}[h]
	\begin{centering}
		\begin{subfigure}[b]{0.238\textwidth}
			\centering
			\includegraphics[width=1\linewidth]{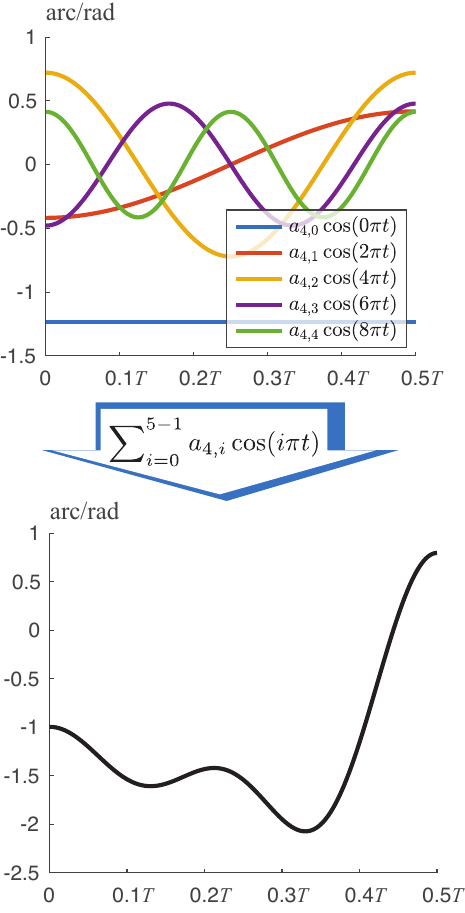}
			\caption{Finite Fourier series}
			\label{fig:ffs}
		\end{subfigure}
		\hfill
		\begin{subfigure}[b]{0.238\textwidth}
			\centering
			\includegraphics[width=1\linewidth]{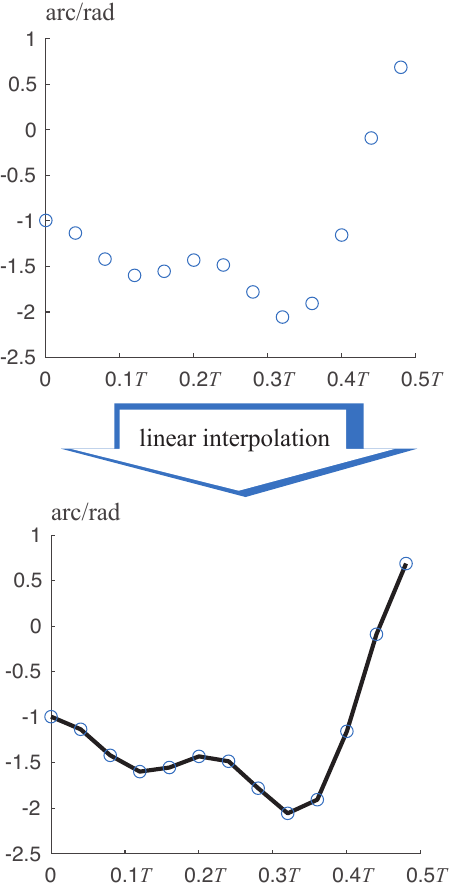}
			\caption{Discrete waypoint series }
			\label{fig:discrete}
		\end{subfigure}
	\end{centering}
	\caption{This figure compares the motions of the $\text{4}^\text{th}$ joint of a manipulator, represented by a combination of a finite Fourier series and by a discrete waypoint series, respectively. 
	\label{fig:ffs-discrete}}
\end{figure}

Unlike some former works that have used a discrete waypoint series to represent the entire trajectory from start $\bm{\theta}_0$ to goal $\bm{\theta}_g$, this paper introduces a novel approach by utilizing the finite Fourier series (FFS) composed of a finite number of harmonics to depict the trajectory in an $M$-dimensional configuration space accurately:
\begin{equation} \label{eq:ffs0}
\begin{aligned}
	\bm{\xi}(t)
	= \left (
	\begin{matrix} 
		a_{1,0} & a_{1,1} & \cdots & a_{1,N} \\
	  	a_{2,0} & a_{2,1} & \cdots & a_{2,N} \\
		\vdots & \vdots & \ddots & \vdots \\
	  	a_{M,0} & a_{M,1} & \cdots & a_{M,N} \\
	 \end{matrix}
	 \right ) 
	 \left (
	 \begin{matrix} 
	  	1 \\
	   	\cos(\frac{2\pi}{T}t) \\
		\vdots \\
		\cos(\frac{2N\pi}{T}t) \\
	\end{matrix}
	\right ) \\
	+ \left (
	\begin{matrix} 
		b_{1,1} & b_{1,2} & \cdots & b_{1,N} \\
	  	b_{2,1} & b_{2,2} & \cdots & b_{2,N} \\
		\vdots & \vdots & \ddots & \vdots \\
	  	b_{M,1} & b_{M,2} & \cdots & b_{M,N} \\
	 \end{matrix}
	 \right ) 
	 \left (
	 \begin{matrix} 
	  	\sin(\frac{2\pi}{T}t) \\
	   	\sin(\frac{4\pi}{T}t) \\
		\vdots \\
		\sin(\frac{2N\pi}{T}t) \\
	  \end{matrix}
	  \right ), 
\end{aligned}
\end{equation}
where $a_{m,n}$ and $b_{m,n}$ denote the amplitude of the harmonics $\cos(\frac{2\pi}{T}nt)$ and $\sin(\frac{2\pi}{T}nt)$ with period $\frac{T}{n}$ of the $m^\text{th}$ joint, respectively, and $\frac{N}{T}$ denotes the maximum discrete frequency. \figref{fig:ffs-discrete} shows how this research aims to overcome certain limitations associated with traditional waypoint-based representations by adopting the FFS method. 

Our former studies \cite{feng2022iSAGO, yichang2021AGP-STO} found that the discrete waypoint series often suffer from uneven distribution and lack of smoothness, leading to suboptimal results in trajectory planning. \figref{fig:stuck-time} shows how the waypoints, denoted by the red balls, collide with the environmental obstacles when the discrete waypoints are smooth and safe. So, we adopted the up-sampling method \cite{Mukadam2018GPMP} to interpolate the mid-waypoints between the support waypoints to ensure the planned motion's time-continuous safety. 
In contrast, the FFS technique offers several advantages. Firstly, it provides a continuous representation of the trajectory by leveraging harmonics \eqref{eq:ffs0}, as shown in \figref{fig:ffs}. It ensures smoother transitions between different points along the path and enables more precise modeling of complex trajectories with varying curvatures. 
Moreover, employing FFS facilitates easier manipulation and analysis of trajectories. The mathematical properties inherent in the Fourier series enable straightforward calculations for various operations in \secref{sec:kinetic} and \secref{sec:potential}, such as interpolation, differentiation, integration, and optimization on trajectories represented in this form. 

\begin{figure}[h]
	\begin{centering}
		\begin{subfigure}[t]{0.232\textwidth}
			\centering
			\includegraphics[width=1\linewidth]{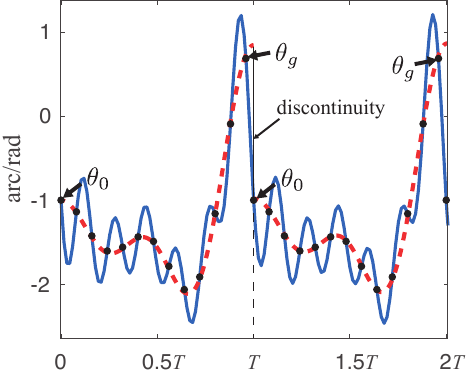}
			\caption{Fourier series}
			\label{fig:dft}
		\end{subfigure}
		\hfill
		\begin{subfigure}[t]{0.242\textwidth}
			\centering
			\includegraphics[width=1\linewidth]{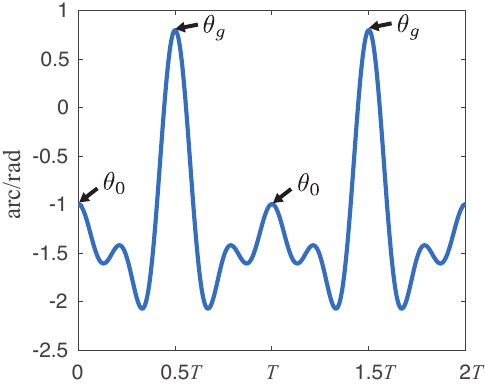}
			\caption{Cosine series}
			\label{fig:dct}
		\end{subfigure}
	\end{centering}
	\caption{This comparison between the $\text{4}^\text{th}$-joint motions, represented by a cosine series and Fourier series, respectively, shows how the overfitting of the discontinuity between $\bm{\xi}(T)$ and $\bm{\xi}(T^+)$ in \figref{fig:dft} brings the extra disturbances into the joint motion.  
	\figref{fig:dct} abridges the original period $T$ adopted by \figref{fig:dft} into $\frac{T}{2}$ and symmetrizes the trajectory according to a dash line $t = \frac{T}{2}$ to smooth the trajectory.   
	\label{fig:dft-dct}}
\end{figure}

However, the FFS representation sometimes encounters the overfitting problem due to the discontinuity at time $\{nT \textbar_{ n = 0,1,2, \dots}\}$, because it views the $T$-long trajectory as the periodic wave with $T$-period. In this way, there exist shifts between $\bm{\xi}(nT) = \bm{\theta}_g$ (goal waypoint) and $\bm{\xi}(nT^{+}) = \bm{\theta}_0$ (start waypoint), where $n$ is an arbitrary integer. \figref{fig:dft} shows how the overfit, and the extra disturbing waves it brings out arise facing the discontinuity.
To eliminate the discontinuity of $\bm{\xi}$, \figref{fig:dct} abridges the motion time into ${T}/{2}$ and symmetrizes the trajectory with respect to $t = {T}/{2}$. Then $\bm{\xi}({T}/{2}) = \bm{\xi}({T^{+}}/{2}) = \bm{\theta}_g$, $\bm{\xi}(T) = \bm{\xi}(T^{+}) = \bm{\theta}_0$ and only the cosine part of the FFS \eqref{eq:ffs0} is conserved due to the trajectory symmetry. 

Overall, this section introduces FFS only with cosine functions for trajectory representation in the high-dimensional configuration space to address the limitations encountered by previous studies, such as time-continuous safety. The following sections will introduce how to calculate the objective functional, a.k.a. Hamiltonian, for motion smoothness and safety.

\section{PROBELM FORMULATION}
\subsection{Kinetic energy}\label{sec:kinetic}
Like the former studies \cite{feng2022iSAGO, yichang2021AGP-STO}, utilizing the dynamical quantities for trajectory smoothness, we first calculate the trajectory velocity according to \eqref{eq:ffs0}: 
\begin{equation}
	\dot{\bm{\xi}} 
	=  \left (
	\begin{matrix} 
		a_{1,1} & a_{1,2} & \cdots & a_{1,N} \\
	  	a_{2,1} & a_{2,2} & \cdots & a_{2,N} \\
		\vdots & \vdots & \ddots & \vdots \\
	  	a_{M,1} & a_{M,2} & \cdots & a_{M,N} \\
	 \end{matrix}
	 \right ) 
	 \left (
	 \begin{matrix} 
	   	\frac{2\pi}{T} \sin(\frac{2\pi}{T}t) \\
		\frac{4\pi}{T} \sin(\frac{4\pi}{T}t) \\
		\vdots \\
		\frac{2N\pi}{T}\sin(\frac{2N\pi}{T}t) \\
	\end{matrix}
	\right ). 
\end{equation}
Then, considering the orthogonality of trigonometric function, we gain the kinetic energy of $\bm{\xi}(t)$ from $t = 0$ to $t = \frac{T}{2}$: 
\begin{equation} \label{eq:kinetic}
	\mathcal{F}_\textit{kinetic} (\bm{\xi}) = \int_{t = 0}^{\frac{T}{2}} \frac{\|\dot{\bm{\xi}}(t)\|^{2}}{2}  \diff t
	=  \frac{\pi}{4} \sum_{m = 1}^{M} \sum_{n = 1}^{N} n^2 a_{m,n}^2. 
\end{equation}
%

\subsection{Potential energy}\label{sec:potential}

Since the collision-free trajectory means a safe motion, this section constructs a collision field, repelling the manipulator out of the collision area based on the closest distance between the manipulator and surrounding obstacles. We first use $\bm{x}_{i,t} = \bm{x}(\mathcal{B}_i, t)$ to map from a state $\bm\theta_t$ at time $t$ to the position state of a collision-check ball (CCB-$\mathcal{B}_i$) on the manipulator. Then we calculate the collision cost 
\begin{equation} \label{eq:collisionCost}
	c(\bm{x}_{i,t}) = \left \{
	\begin{array}{cl}
		0 & d(\bm{x}_{i,t}) > \epsilon \\
		\epsilon - d(\bm{x}_{i,t}) & d(\bm{x}_{i,t}) \leq \epsilon \\
	\end{array} \right.
\end{equation}
which increases when the signed distance $d(\bm{x}_{i,\tau})$ between $\mathcal{B}_i$ and its closest obstacle decreases and uses $\epsilon$ to define a buffer zone between the safe and unsafe areas. Then, we adopt the obstacle functional proposed by \cite{} to estimate the potential energy: 
\begin{equation}\label{eq:potential}
	\mathcal{F}_\textit{obs} (\bm{\xi}) = \int_{t = 0}^{\frac{T}{2}} \max_{\mathcal{B}_i \in \bm{\mathcal{B}}} c(\bm{x}_{i,t}) \|\bm{x}_{i,t}\| \diff t
\end{equation}
where $\bm{\mathcal{B}} = \{\mathcal{B}_1, \dots,\mathcal{B}_{|\bm{\mathcal{B}}|}\}$ consists of all collision-check balls of a robot. 

Though our FFS method \eqref{eq:ffs0} combines a series of cosine functions to represent the time-continuous trajectory, we still could not find a way to represent the $\bm{x}(t)$ analytically in a continuous form. Therefore, considering the linearization and quasi-Hessian estimation, we first discretize the trajectory $\bm{\xi}$. Since the discrete trajectory composes of a series of waypoints $\hat{\bm{\xi}} = [\bm{\theta}_0^\mathrm{T}, \bm{\theta}_1^\mathrm{T}, \cdots, \bm{\theta}_{T/2}^\mathrm{T}]^\mathrm{T}$, we transform the amplitude matrix of \eqref{eq:ffs0} into a $M(N+1)$-dimensional vector
$$\bm{a} = [a_{1,0} \cdots a_{1,N},~ a_{2,0} \cdots a_{2,N}, \cdots, a_{M,0} \cdots a_{M,N}]^\mathrm{T}$$
and use the tensor product to denote the waypoint
\begin{equation} \label{eq:singleWaypoint}
	\bm{\theta}_{t} = \mathbf{C}_{t} \bm{a}, 
	\mathbf{C}_t = \bm{\mathbf{I}} \otimes \left(1, \cos(\tfrac{2\pi}{T}t), \cdots, \cos(\tfrac{2N\pi}{T}t) \right), 
\end{equation}
where $\mathbf{I}$ is an $M \times M$ identical matrix. In this way, the discrete trajectory can be transformed as
\begin{equation}\label{eq:ffs-discrete}
	\hat{\bm{\xi}} = \bm{\mathcal{C}} \cdot \bm{a},
\end{equation}
where $\bm{\mathcal{C}} = [\mathbf{C}_0^\mathrm{T}, \mathbf{C}_1^\mathrm{T}, \cdots,  \mathbf{C}_{T/2}^\mathrm{T}]^{\mathrm{T}}$. 

 Then we adopt the time-discrete form proposed by \cite{Mukadam2018GPMP} to estimate its potential energy 
\begin{equation}\label{eq:Fp}
	\mathcal{F}_\textit{potential} (\hat{\bm{\xi}}) = \sum_{t = 0}^{T/2} f_{p}[\bm{\theta}_t]^2, 
\end{equation}
\begin{equation}\label{eq:fp}
	f_{p}(\bm{\theta}_t) = \max_{\mathcal{B}_i \in \bm{\mathcal{B}}} c\left[\bm{x}_{i}(\mathbf{C}_{t}\bm{a})\right] \cdot \|\bm{x}_{i}(\mathbf{C}_{t}\bm{a})\|.
\end{equation}

According to the derivation of the functional gradient in \cite{Zucker2013CHOMP}, we get the gradient of $f_p$
\begin{equation}\label{eq:grad_fp}
	\nabla_{\bm{a}} f_{p}(\bm{\theta}_t) = \tfrac{1}{N}\mathbf{C}_{t}^\mathrm{T}\max_{\mathcal{B}_i \in \bm{\mathcal{B}}} \bm{J}^{\mathrm{T}}\left\|\dot{\bm x}_i\right\|[(\mathbf{I}-\hat{\dot{\bm x}}_i \hat{\dot{\bm x}}_i^{\mathrm{T}}) \nabla c - c \bm\kappa]
\end{equation}
where $\bm{J} = \nabla_{\bm{\theta}} \bm x_i$ is the kinematic Jacobian, $\dot{\bm x}_i = \bm{J} \dot{\mathbf{C}}_t \bm{a}$ calculates the velocity of $\mathcal{B}_i$, $\hat{\dot{\bm x}}_i = \frac{\dot{\bm x}_i}{\|\dot{\bm x}_i\|}$ is the normalized velocity, $\ddot{\bm x}_i \approx \bm{J} \ddot{\mathbf{C}}_t \bm{a}$ calculates the acceleration of $\mathcal{B}_i$, $\bm\kappa=\left\|\dot{\bm x}_i\right\|^{-2}\left(\mathbf{I} - \dot{\bm x}_i \dot{\bm x}_i^{\mathrm{T}}\right) \ddot{\bm x}_i$, and $\mathbf{I}$ is a $3\times 3$ identical matrix. 
\secref{sec:AIP} will use the above gradient information to approximate $\mathcal{F}_\textit{potential}$ in a second-order nonlinear form. 

Though this section adopts the discrete form \eqref{eq:ffs-discrete} for numerical calculation \eqref{eq:Fp} of the potential energy in continuous time, the stuck cases in the time scale can still be overcome since this paper optimizes the trajectory by the harmonic adjustment based on the $\bm{a}$-value \eqref{eq:grad_fp} rather than by the discrete waypoint adjustment. In this way, our method does not require the mid-waypoint interpolation between the two adjacent support waypoints to guarantee time-continuous safety, reducing the number of waypoints and improving computational efficiency. The results in \tabref{table:results_C:iiwa} validate the FFS' ability of stuck overcoming in the time scale and its efficiency. 

\subsection{Hamiltonian construction}\label{sec:objective}
\secref{sec:kinetic} and \secref{sec:potential} formulate the kinetic and potential energies, respectively. The kinetic part \eqref{eq:kinetic} tends to smooth the trajectory, whereas the potential part drives the trajectory away from the obstacles. In this way, this paper introduces a factor $\varrho$ to balance between the smoothness and safety and designs the Hamiltonian similar to \cite{Zucker2013CHOMP, Mukadam2018GPMP}
\begin{equation}
	\mathcal{F}(\bm{\xi}) = \varrho \mathcal{F}_\textit{kinetic}(\bm{\xi}) + \mathcal{F}_\textit{potential}(\bm{\xi}).
\end{equation}
So we can approach the optimal trajectory following the descent direction of the functional gradient $\bar{\nabla}\mathcal{F}(\bm{\xi})$ until the functional extreme value is found, i.e. $\bar{\nabla}\mathcal{F}(\bm{\xi}) = 0$.

\subsection{Constraint construction}\label{sec:constraint}

\subsubsection{Start and goal constraints}
Since the planning problem this paper tends to solve predefines the start and goal waypoints $\bm{\theta}_0, \bm{\theta}_{g}$, we can gain the two equality constraints 
\begin{align}
	\mathbf{C}_{0} \bm{a} &= \bm{\theta}_{0}, \\
	\mathbf{C}_{{T}/{2}} \bm{a} &= \bm{\theta}_{g}. 
\end{align}
to satisfy the start and goal conditions according to \eqref{eq:singleWaypoint}. 

\subsubsection{Joint limit constraints}
Besides the collision avoidance, we also need to bound the joint state inside the available range to ensure the motion safety. To ensure the trajectory $\bm{\xi}$ in \eqref{eq:ffs0} within the joint limits $[\bm{\theta}_{\min}, \bm{\theta}_{\max}]$, we only need to ensure the maximum and minimum values of $\bm{\xi}$ under these constraints. So we find the waypoints $\bm{\theta}_t$ of $\bm{\xi}$ satisfying $\dot{\bm{\xi}}(t) = 0$ and then use the maximum and minimum among them. However, solving the nonlinear equation $\dot{\bm{\xi}}(t) = 0$ is time consuming. So we only check the discrete trajectory $\hat{\bm{\xi}} = \bm{\mathcal{C}}\cdot\bm{a}$ in \eqref{eq:ffs-discrete} and define its inequality constraints as
\begin{align} \label{eq:jointLimits}
	\bm{\mathcal{C}}\cdot \bm{a} &\leq \mathbf{e} \otimes \bm{\theta}_{\max}, \\
	-\bm{\mathcal{C}}\cdot \bm{a} &\leq -\mathbf{e} \otimes \bm{\theta}_{\min}, 
\end{align}
where $\mathbf{e} = [1,..., 1]^\mathrm{T}$ is a $(T+1)$-dimensional vector, to ensure that a robot with $(T+1)$ waypoints moves within the joint limits $[\bm{\theta}_{\min}, \bm{\theta}_{\max}]$. 

\section{METHODOLOGY}

\subsection{Adaptive interior point optimization}\label{sec:AIP}

Like some former studies \cite{Zucker2013CHOMP, Mukadam2018GPMP, Bhardwaj2020dGPMP, Schulman2013SCO} utilizes the numerical optimization methods to find a local minimum of the objective functional, this paper implements the interior point methods to solve the objective, i.e. Hamiltonian, defined in \secref{sec:objective}, under the constraints, defined in \secref{sec:constraint}: 
\begin{equation}
\begin{aligned}
	&\argmin_{\bm{a}} \mathcal{F}(\bm{a}) \\	
	\text{s.t.}    
	& \left[
	\begin{matrix} 
	      	\mathbf{C}_0 \\
	      	\mathbf{C}_{\frac{T}{2}} \\
	\end{matrix}
	\right]
	\bm{a}
	= \left[
	\begin{matrix} 
		\bm{\theta}_0 \\
	      	\bm{\theta}_{g} \\
	\end{matrix}
	\right], \\
	& \left[
	\begin{matrix} 
	      	\bm{\mathcal{C}} \\
	      	-\bm{\mathcal{C}} \\
	\end{matrix}
	\right]
	\bm{a}
	\leq \left[
	\begin{matrix} 
		\bm{e} \otimes \bm{\theta}_{\max} \\
	      	-\bm{e} \otimes \bm{\theta}_{\min} \\
	\end{matrix}
	\right]. \\
\end{aligned}	
\end{equation}

The former study \cite{Schulman2013SCO} uses $\ell_1$-penalty augmentation and trust-region method, and some studies \cite{Zucker2013CHOMP, Mukadam2018GPMP} use line search methods for constrained trajectory optimization, this paper adopts the interior-point method proceeding the prime-dual steps to satisfy the KKT condition \cite{Nocedal2006NumericalOpt}. So the objective $\mathcal{F}(\bm{a})$ must have a quadratic form, and its Hessian matrix must be positive definite.  
The convexity of the kinetic functional 
\begin{equation}
	\mathcal{F}_\textit{kinetic} = \bm{a}^\mathrm{T} \mathbf{K} \bm{a} 
\end{equation}
is attributed to its constant positive-definite Hessian matrix 
$$\mathbf{K} = \mathbf{I} \otimes \text{diag}(N^2, (N-1)^2, \cdots, 1, 0)$$ according to \eqref{eq:kinetic}. 
However, when considering the overall objective, it becomes apparent that the non-convexity primarily arises from the potential part. Due to its inherent nature, this component introduces additional complexities and variations in the objective function.
To solve this concern, we utilize the functional gradient of $\mathcal{F}_\textit{potential} (\bm{\xi})$ \eqref{eq:grad_fp} and approximated it in a quadratic form as
\begin{equation}
\begin{aligned}
	\hat{\mathcal{F}}_\textit{potential} (\bm{\xi}) &=  \sum_{t = 0}^{T/2} \left[f_{p}(\bm{\theta}_t) + \nabla_{\bm{a}} f_{p}(\bm{\theta}_t) \delta \bm{a}\right]^{2} \\
	& = \|\mathbf{f} + \mathbf{F}\delta\bm{a}\|^2 + \delta\bm{a}^\mathrm{T}\bm{\Lambda}\delta\bm{a},  
\end{aligned}
\end{equation}
where $$\mathbf{f} = [f_p(\bm{\theta}_0), f_p(\bm{\theta}_1), \dots, f_p(\bm{\theta}_{\frac{T}{2}})]$$ is a $({T}/{2} + 1)$-dimensional vector consisting of a series of potential powers from time $0$ to $T/2$, 
$$\mathbf{F} = [\nabla_{\bm{a}}f_p(\bm{\theta}_0), \nabla_{\bm{a}}f_p(\bm{\theta}_1), \dots, \nabla_{\bm{a}}f_p(\bm{\theta}_{\frac{T}{2}})]^\mathrm{T},$$ a $M(N+1)\times ({T}/{2} + 1)$ matrix, consists of the functional gradients of $\mathcal{F}_\textit{potential}(\hat{\bm{\xi}})$, 
and $\mathbf{\Lambda}$ is a scaled $M(N+1)\times M(N+1)$ identical matrix which refines the gradient step-size repressing the overstrike of the objective value. 
Then we can gain the approximated functional objective as
\begin{equation}
\begin{aligned}
	\hat{\mathcal{F}}(\delta \bm{a}) = & \delta\bm{a}^\mathrm{T} (\varrho \mathbf{K} + \mathbf{F}^\mathrm{T}\mathbf{F} + \mathbf{\Lambda}) \delta\bm{a} \\
	&+ 2(\varrho\bm{a}^\mathrm{T}\mathbf{K} + \mathbf{f}^\mathrm{T} \mathbf{F})\delta\bm{a}. \\
\end{aligned}
\end{equation}
In this way, the sequential quadratic programming can be applied to solve the above constrained quadratic problem iteratively until $\nabla \hat{\mathcal{F}}_\textit{potential} = 0$. 

However, the above method sometimes introduces some vibrations into the optimization. It can be attributed to the interplay between the potential functional $\mathcal{F}_\textit{potential}$ and kinetic functional $\mathcal{F}_\textit{kinetic}$ during the optimization. When applying the potential functional $\mathcal{F}_\textit{potential}$, it forces the manipulator out of the collision areas. However, once the manipulator is successfully pushed into a safe area, a shift occurs where the kinetic functional $\mathcal{F}_\textit{kinetic}$ takes over as the main driving force and may push the trajectory back to the collision areas. That is because our method sets a constant $\mathbf{\Lambda}$ to save time for its value search and sometimes cannot adjust the step size. 
In this transition from potential to kinetic, there may be instances where vibrations are introduced due to conflicting forces acting on the manipulator. In this way, \algref{alg:AIP} adopts the adaptive gradient method \cite{Kingma2014Adam} and updates exponential moving averages of the gradients
\begin{align}
	\mathbf{f}_i^\textit{history} &= (1-\alpha)\mathbf{f}_{i-1}^\textit{history} + \alpha \mathbf{f}_i, \\
	\mathbf{F}_i^\textit{history} &= (1-\beta)\mathbf{F}_{i-1}^\textit{history} + \beta \mathbf{F}_i,
\end{align}
where the hyper-parameters $\alpha, \beta$ control the exponential decay rates of these moving averages. This approach is widely used in optimization algorithms to improve convergence speed and stability. By updating the moving averages, we can effectively capture both short-term fluctuations and long-term trends in the gradients, allowing us to make more informed decisions during optimization. Since there exists some bias between the moving average and the moving expectation according to \cite{Kingma2014Adam}, we correct the bias by
\begin{align}
	\hat{\mathbf{f}}_i = \frac{\mathbf{f}_i^\textit{history}}{1-(1-\alpha)^i}, 
	\hat{\mathbf{F}}_i = \frac{\mathbf{F}_i^\textit{history}}{1-(1-\beta)^i}. 
\end{align}
Finally, we implement the adaptive interior-point method (AIP, \algref{alg:AIP}) to solve the above problem iteratively until the $\delta\bm{a}$ meets the step-size threshold \textit{stepTol}. 
\begin{algorithm}[htbp]
\caption{AIP}\label{alg:AIP}
\DontPrintSemicolon
\LinesNumbered
\SetKwInOut{Input}{Input}
\SetKwInOut{Output}{Output}
\SetKwFunction{Union}{Union}
\Input {start and goal waypoints $\bm{\theta}_0, \bm{\theta}_g$, functionals $\mathcal{F}_\textit{kinetic}, \mathcal{F}_\textit{potential}$.} 
\Output {optimized $\bm{a}$. }
\textbf{Initialization:} $\mathbf{f}^\textit{history} = \mathbf{0}, \mathbf{F}^\textit{history} = \mathbf{O}$, 
\begin{align*}
\bm{a}_0 &= \argmin_{\bm{a}} \mathcal{F}_\textit{kinetic}(\bm{a})\\
& \text{s.t.} 
	\left[
	\begin{matrix} 
	      	\mathbf{C}_0 \\
	      	\mathbf{C}_{\frac{T}{2}} \\
	\end{matrix}
	\right] \bm{a}
	= \left[
	\begin{matrix} 
		\bm{\theta}_0 \\
	      	\bm{\theta}_{g} \\
	\end{matrix}
	\right]
\end{align*} \\

\For(\tcp*[h]{\scriptsize optimize via adaptive interior-point optimization}){$ \textit{AIPIter}: i = 1 \dots N_\text{aip} $}{
	Calculate $\mathbf{f}(\bm{a}), \mathbf{F}(\bm{a})$; \;
	Update $\mathbf{f}^\textit{history} \leftarrow (1-\alpha)\mathbf{f}^\textit{history} + \alpha \mathbf{f}$; \;
	Update $\mathbf{F}^\textit{history} \leftarrow (1-\beta)\mathbf{F}^\textit{history} + \beta \mathbf{F}$; \;
	Bias correction $\hat{\mathbf{f}} = \frac{\mathbf{f}^\textit{history}}{1-(1-\alpha)^i}, 
	\hat{\mathbf{F}} = \frac{\mathbf{F}^\textit{history}}{1-(1-\beta)^i}; $ \;
	$\hat{\mathcal{F}} \triangleq \delta\bm{a}^\mathrm{T} (\varrho \mathbf{K} + \hat{\mathbf{F}}^\mathrm{T}\hat{\mathbf{F}} + \mathbf{\Lambda}) \delta\bm{a} 
		 + 2(\varrho\bm{a}^\mathrm{T}\mathbf{K} +  \hat{\mathbf{f}}^\mathrm{T} \hat{\mathbf{F}})\delta\bm{a} $; \\
	$\delta \bm{\alpha} = \argmin_{\delta\bm{a}} \hat{\mathcal{F}} $ by interior-point algorithm
	\begin{align*}
		 \text{s.t.}    
	& \left[
	\begin{matrix} 
	      	\mathbf{C}_0 \\
	      	\mathbf{C}_{{T}/{2}} \\
	\end{matrix}
	\right] \delta \bm{a}
	= \left[
	\begin{matrix} 
		\bm{\theta}_0 - \mathbf{C}_0 \bm{a} \\
	      	\bm{\theta}_{g} - \mathbf{C}_{{T}/{2}} \bm{a} \\
	\end{matrix}
	\right], \\
	&\left[
	\begin{matrix} 
	      	\bm{\mathcal{C}} \\
	      	-\bm{\mathcal{C}} \\
	\end{matrix}
	\right]\delta\bm{a}
	\leq \left[
	\begin{matrix} 
		\bm{e} \otimes \bm{\theta}_{\max} - \bm{\mathcal{C}}\bm{a} \\
	      	-\bm{e} \otimes \bm{\theta}_{\min} + \bm{\mathcal{C}} \bm{a}\\
	\end{matrix}
	\right];
	\end{align*}\\
	$\bm{a} \leftarrow \bm{a} + \delta \bm{a}$; \;
	
	\lIf{$\|\bm{a}\| \leq \text{stepTol}$}{
		\Return. 
	}
}

\end{algorithm}

However, despite the advancements made by AIP, it still faces challenges when solving the stuck-case in the space scale, as shown in \figref{fig:stuckCase}. It is primarily due to the non-convex nature of the collision cost associated with a single waypoint. So, the following section proposes an alternative approach using support vector machine (SVM) technology to overcome this limitation.

\subsection{Learn the collision field by SVM} \label{sec:SVM}

By leveraging SVM, we aim to learn a convex collision field that can effectively address the stuck-case problem. SVM is a robust machine learning algorithm that has been widely used in various fields, such as image recognition and natural language processing. Its ability to handle nonlinear data and find optimal separating hyperplanes makes it particularly suitable for our task. Unlike the iSVM \cite{Das2020iSVM} learning in the high-dimensional configuration, this section adopts the method proposed in \cite{Mirrazavi2018unifiedFramework} and learns the hyperplane between the safe and unsafe areas in the Euclidean space. 



\subsubsection{Dataset construction} 

\begin{figure}[htb]
	\begin{centering}
		\includegraphics[width=.75\linewidth]{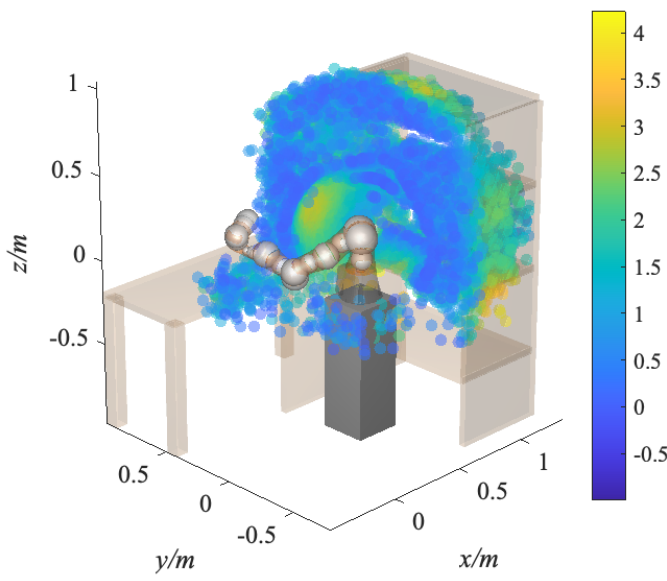}
		\caption{Support vector machine}
		\label{fig:svm}
	\end{centering}
\protect\caption{
The yellow points denote the support vectors forming the hyperplane between the collided and safe areas in the 3D-Euclidean space. The scattering of the support points demonstrates the out layer of the bookshelf is the useless information which traps the manipulator. 
\label{fig:SVM}}
\end{figure}

To learn the collision field $c(\bm{x})$ in \eqref{eq:collisionCost}, it is crucial for us first to generate a comprehensive data set that can accurately identify the boundaries of collisions. To achieve this, we first utilize this simplified geometric model $\bm{\mathcal{B}} = \{\mathcal{B}_1, \dots,\mathcal{B}_{|\bm{\mathcal{B}}|}\}$, a geometrical combination of collision-check balls (CCBs) shown in \figref{fig:SVM}. Then we generate $N_\text{data} = \text{1e5}$ points obeying the uniform distribution $\mathcal{U}(\bm{\theta}_{\min}, \bm{\theta}_{\max})$ and label them with either 'collided' or 'safe' tag according to the signed distance $d(\bm{x}_{n})$ introduced in \secref{sec:potential}: 
\begin{equation} \label{eq:collisionTag}
	y(\bm{x}_{n}) = \left \{
	\begin{array}{ll}
		-1, \text{safe} & d(\bm{x}_{n}) > \epsilon \\
		+1, \text{collided} & d(\bm{x}_{n}) \leq \epsilon. \\
	\end{array} \right.
\end{equation}
%

\subsubsection{Learning-based collision field} 

Considering the nonlinear collision field in 3D-Euclidean space, we employ the Gauss kernel with a scalar $\sigma$
\begin{equation}
	k\left(\bm{x}(\bm{\theta}_i), \bm{x}(\bm{\theta}_j)\right)=\exp{\left(-\frac{1}{2 \sigma^2}\left\|\bm{x}(\bm{\theta}_i)-\bm{x}(\bm{\theta}_j)\right\|^2\right)}
\end{equation}
to effectively learn and model the hyperplane that separates the safe regions from the collided areas. 
In this way, we can estimate the label of the training data $\bm{\theta}_i$ by
\begin{equation}
	\hat{y}_i  = \sum_{n=1}^{N_\text{sv}} \alpha_n y_n k\left(\bm{x}(\bm{\theta}_i), \bm{x}(\bm{\theta}_n)\right)+b, 
\end{equation}
where $\alpha_n$, a non-negative factor combined with a support vector $\bm{x}(\bm{\theta}_n)$ labelled by $y_n$, determines the shape of the hyperplane, while
\begin{equation*}
	b = -\frac{1}{2} \sum_{n=1}^{N_\text{sv}} \alpha_n y_n k\left(\bm{x}(\bm{\theta}_\text{collided}), \bm{x}(\bm{\theta}_n) \right) + k\left(\bm{x}(\bm{\theta}_\text{safe}), \bm{x}(\bm{\theta}_n)\right)
\end{equation*}
determines the position of the hyperplane, calculated by any collided and safe points $\bm{\theta}_\text{collided}, \bm{\theta}_\text{safe}$ from the dataset. Then we employ the sequential minimal optimization (SMO) to gain the optimal $\alpha_n$, satisfying $\alpha_n(\hat{y}_i {y}_i - 1) = 0$. 
Sequentially, we can learn the convex collision field expressed by a linear combination of $N_\text{sv}$ kernel functions shaped by the support vectors: 
\begin{equation}
	\hat{c}\left(\bm{x}(\bm{\theta})\right)  = \max\left(\sum\nolimits_{n=1}^{N_\text{sv}} \alpha_n y_n k\left(\bm{x}(\bm{\theta}), \bm{x}(\bm{\theta}_n)\right)+b+1, 0 \right). 
\end{equation}
as well as its gradient
\begin{equation}
	\nabla \hat{c}\left(\bm{x}(\bm{\theta})\right) = \sum_{n=1}^{N_\text{sv}}-\frac{\alpha_n y_n}{\sigma^2}  k\left(\bm{x}(\bm{\theta}), \bm{x}(\bm{\theta}_n)\right) \left(\bm{x}(\bm{\theta})- \bm{x}(\bm{\theta}_n)\right). 
\end{equation}

Now we define any $\bm{x}_1, \bm{x}_2$ in the 3D Euclidean space and a scalar $\gamma \in [0,1]$ and get 
\begin{equation}
\begin{aligned}
& \gamma \exp(\|\bm{x}_1 - \bm{x}(\bm{\theta}_n)\|^2 + (1-\gamma)\exp\|\bm{x}_2 - \bm{x}(\bm{\theta}_n)\|^2) \\
& \geq 2\sqrt{\gamma(1-\gamma)} \exp \left(\frac{\|\bm{x}_1 - \bm{x}(\bm{\theta}_n)\|^2 + \|\bm{x}_2 - \bm{x}(\bm{\theta}_n)\|^2}{2} \right) \\
& \geq \exp(\gamma\|\bm{x}_1 - \bm{x}(\bm{\theta}_n)\|^2 + (1-\gamma) \|\bm{x}_2 - \bm{x}(\bm{\theta}_n)\|^2)\\
& \geq \exp(\|\gamma[\bm{x}_1 - \bm{x}(\bm{\theta}_n)] + (1-\gamma) [\bm{x}_2 - \bm{x}(\bm{\theta}_n)]\|^2)\\
& = \exp (\|\gamma\bm{x}_1 + (1-\gamma)\bm{x}_2 - \bm{x}(\bm{\theta}_n)\|^2), \\ 
\end{aligned}
\end{equation}
which means all kernels $\{k(\bm{x}, \bm{x}(\bm{\theta}_n))\textbar_{n = 1,\dots,N_\text{sv}} \}$ are convex due to \cite{Boyd2004ConvexOpt}. Therefore, the learning-based collision field $\hat{c}(\bm{x})$ in \figref{fig:SVM}, a linear combination of these convex kernels, is convex in the Euclidean space.  

So we can implement this learning-based collision field $\hat{c}(\bm{x})$ into potential energy \eqref{eq:fp} and its functional gradient \eqref{eq:grad_fp}.
The utilization of the Gauss kernel allows us to accurately represent intricate collision patterns that may arise due to various factors such as obstacle shapes, postures and positions. By leveraging this sophisticated approach, we can overcome limitations posed by linear models when dealing with nonlinearity in collision detection. 

\section{EXPERIMENT}

This section will detail the benchmark and parameter setting in Sections~\ref{sec:setup}~\&~\ref{sec:parameters} and analyze the benchmark results in Section~\ref{sec:analysis}. 

\subsection{Evaluation} \label{sec:Evaluation}

\subsubsection{Setup for benchmark}\label{sec:setup}
This paper benchmarks \ffsomp against the numerical planners (CHOMP~\cite{Zucker2013CHOMP}, TrajOpt~\cite{Schulman2013SCO}, GPMP2~\cite{Mukadam2018GPMP}) and the sampling planners (STOMP~\cite{Kalakrishnan2011STOMP}, RRT-Connect~\cite{Kuffner2000RRT-connect}) on  a 7-DoF robot (LBR-iiwa)  and a 6-DoF robot (AUBO-i5). Since the benchmark executes in MATLAB, we use \code{BatchTrajOptimize3DArm} of GPMP2-toolbox to implement GPMP2, \code{plannarBiRRT} with 20s maximum time for RRT-Connect, \code{fmincon}\footnote{We use \code{optimoptions('fmincon','Algorithm','trust\--region\--reflective','Specify\-Objective\-Gradient', true)} to apply \code{fmincon} for the trust-region method like TrajOpt and calculates \code{Objective\-Gradient} analytically and \code{Aeq} (a matrix whose rows are the constraint gradients) by numerical differentiation.} for TrajOpt, \code{hmcSampler}\footnote{Our benchmark defines an \code{hmcSampler} object whose logarithm probabilistic density function \code{logpdf} is defined by \cite{Zucker2013CHOMP}, uses \code{hmcSampler.drawSamples} for HMC adopted by CHOMP. } for CHOMP, and \code{mvnrnd}\footnote{STOMP samples the noise trajectories by \code{mvnrnd} and updates the trajectory via projected weighted averaging. } for STOMP. Since all of them are highly tuned in their own studies, our benchmark uses their default settings. 

\begin{figure}[h]
	\begin{centering}
		\begin{subfigure}[b]{0.23\textwidth}
			\centering
			\includegraphics[width=1\linewidth]{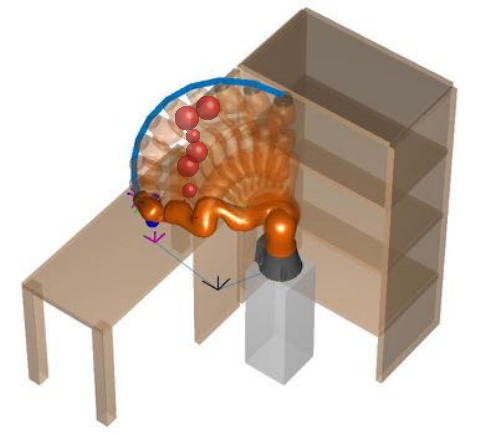}
			\caption{Task A-1, 6 red CCBs}
			\label{fig:iiwa:A1}
		\end{subfigure}
		\hfill
		\begin{subfigure}[b]{0.23\textwidth}
			\centering
			\includegraphics[width=1\linewidth]{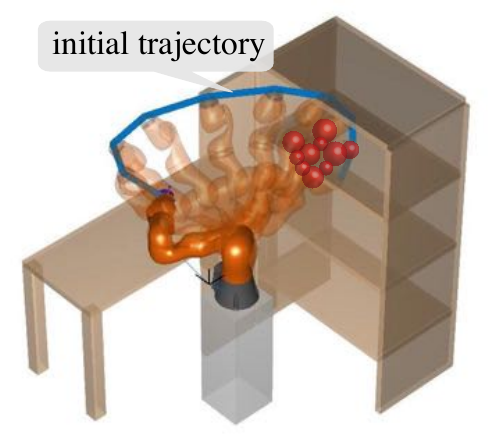}
			\caption{Task B-1, 11 red CCBs}
		\end{subfigure}
		\hfill
		\begin{subfigure}[b]{0.23\textwidth}
			\centering
			\includegraphics[width=1\linewidth]{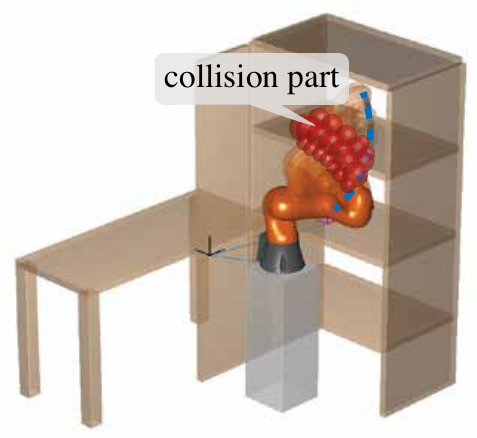}
			\caption{Task C-3, 22 red CCBs}
			\label{fig:iiwa:C3}
		\end{subfigure}
		\hfill
		\begin{subfigure}[b]{0.23\textwidth}
			\centering
			\includegraphics[width=1\linewidth]{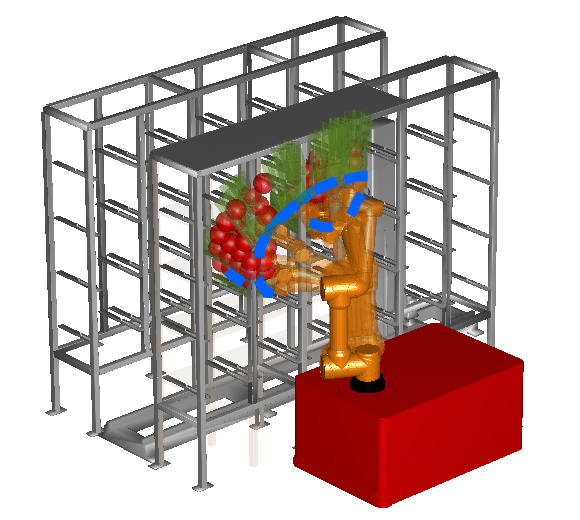}
			\caption{Task C-6, 35 red CCBs}
			\label{fig:AUBO:C6}
		\end{subfigure}
	\end{centering}
	\caption{The initial trajectory with red collision parts visualizes our benchmark on LBR-iiwa or AUBO-i5. \textit{Task C-6} means the No.{6} task of class {C}. 
	\label{fig:problems_ABC}}
\end{figure}

To illustrate the competence of \ffsomp for planning tasks, we conduct 25 experiments on LBR-iiwa at a bookshelf and 12 experiments on AUBO-i5 at a storage shelf. We categorize all tasks into 3 classes (A, B, and C) whose planning difficulty rises with the stuck cases in the initial trajectory increase. Figure~\ref{fig:problems_ABC} visually validates our classification because the number of red in-collision CCBs increases from Task A-1 to Task C-3. Considering the difficulty of different classes, we first generate 2 tasks of class A, 3 tasks of class B, and 4 tasks of class C for LBR-iiwa (Figures~\ref{fig:iiwa:A1}-\ref{fig:iiwa:C3}). Then we generate 6 tasks of class C for AUBO-i5 (Figures~\ref{fig:AUBO:C6}). Each task consists of 2 to 4 problems\footnote{Since LBR-iiwa has 7 DoFs, meaning infinite solutions for the same goal constraint, our benchmark uses LM~\cite{Levenberg1944LM} with random initial points to generate different goal states. So one planning task of LBR-iiwa with the same goal constraint has several problems with different goal states. Meanwhile, each task of AUBO-i5 has two problems with the same initial state and goal constraint because the same goal constraint has only one solution for the 6-DoF, and we switch yaw-angle in $\{0\degree, 180\degree\}$ for each. } with the same initial state and goal constraint. Moreover, we compare \ffsomp with AIP to show the efficiency and reliability of implementing the SVM-based learning procedure. 
\begin{figure}[htb]
	\begin{centering}
		\begin{subfigure}[b]{0.24\textwidth}
			\centering
			\includegraphics[width=1\linewidth]{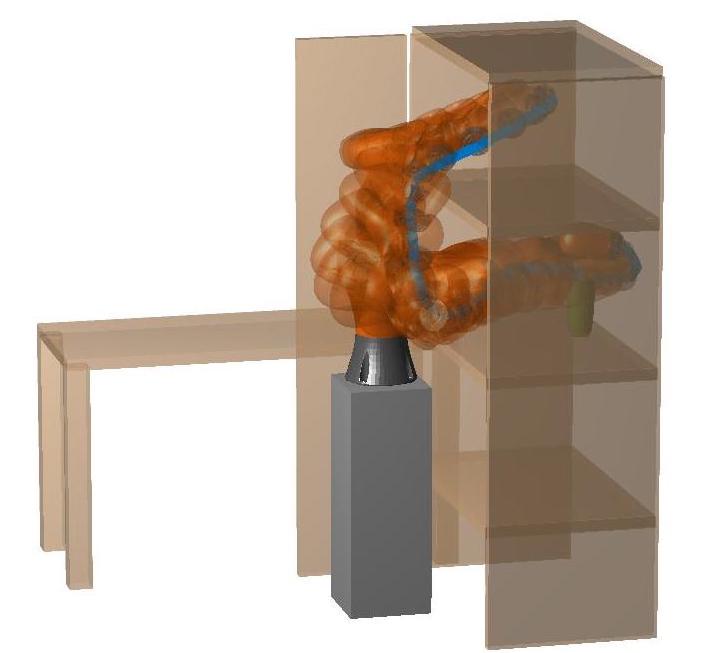}
			\caption{Trajectory C-3.2}
			\label{fig:iiwa:C3.2}
		\end{subfigure}
		\hfill
		\begin{subfigure}[b]{0.22\textwidth}
			\centering
			\includegraphics[width=1\linewidth]{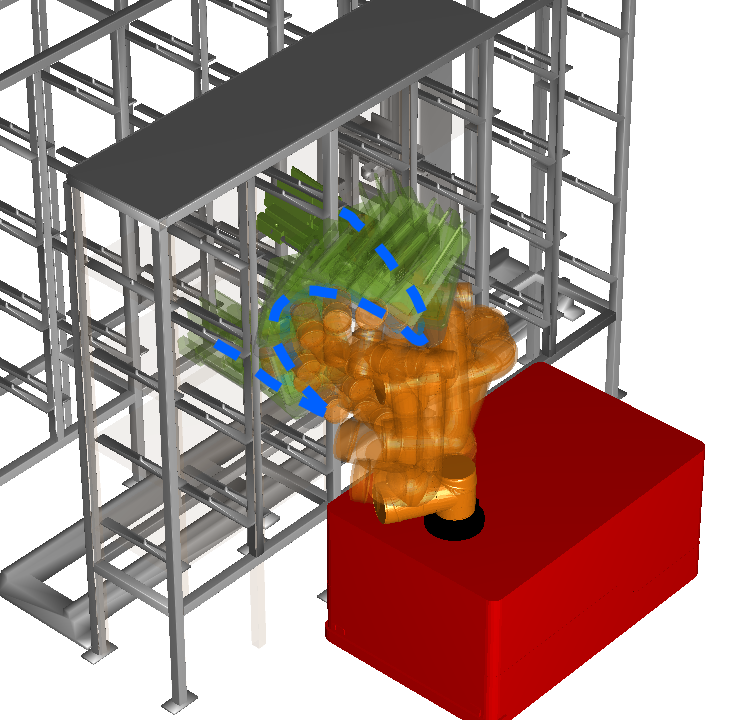}
			\caption{Trajectory C-6.2}
			\label{fig:AUBO:C6.2}
		\end{subfigure}	
		\hfill
	\end{centering}
	\caption{Some results of \ffsomp in class C  (\textit{Trajectory C-3.2} means the 2nd planning result of task 3 in class C. )  
	\label{fig:results_C}}
\end{figure}
%

\subsubsection{Parameter setting}\label{sec:parameters}

Since AIP (Algorithm~\ref{alg:AIP}) proceeds sequential constrained optimization roughly contained in a $\|\mathbf{\Lambda}\|_\infty$-sized region with $\alpha/\beta$-damped EMA of historical momenta, this section first tunes these two parameters, then the others.

According to abundant experiences, we tune the key parameters $\|\mathbf{\Lambda}\|_\infty \in [\text{1e-3}, 1]$, $\alpha,\beta \in [0.50, 0.99]$ and conduct 10 different class C tasks for each pair of parameters. Table~\ref{table:results_A} indicates that the convergence rate increases with $\|\mathbf{\Lambda}\|_\infty$ expansion. In contrast, the success rate decreases with the excessive $\|\mathbf{\Lambda}\|_\infty$ (i.e., too high or too low). Moreover, the SGD rate increases with $\alpha,\beta$-shrink while the success rate decreases with $\alpha,\beta$-shrink. All in all, \ffsomp performs more stable with a smaller step-size, i.e. larger $\|\mathbf{\Lambda}\|_\infty$, hampering local minima's overcoming. Meanwhile, it combines less historical momenta to perform faster, reducing the reliability. So we set $\|\mathbf{\Lambda}\|_\infty = 1e-2$ and $\alpha, \beta = 0.90$.  
\begin{table}
\caption{Results for $\{\alpha, \beta\}$ tuning of 10 tasks in class \textbf{C}}
\label{table:results_A}
\centering
\begin{threeparttable}
{
\begin{tabular}{l|ccccccc}
\toprule
\scalebox{0.8}{\diagbox{$\alpha, \beta$}{(\%)|(s)}{$\|\bm{\Lambda}\|$}} & 1e-3 & 1e-2 & 1e-1 & 1 \\ 
\midrule
 0.50 & 80 | 1.853 & 90 | 2.098 & 60 | 4.398 & 40 | 7.359 \\
 0.90 & 90 | 2.712 & \bf{100} | \bf{2.263} & 80 | 6.033 & 60 | 9.019 \\
 0.95 & 90 | {2.692} & \bf{100} \textnormal{| 4.732} & 80 | 7.931 & 60 | 13.63 \\
\bottomrule
\end{tabular}}
\begin{tablenotes}\scriptsize
     \item[1] (\%) and (s) denote the success rate and average computation time. 
\end{tablenotes}
\end{threeparttable}
\end{table}
%
 

\subsubsection{Result analysis}\label{sec:analysis}

\begin{table*}[htp]
\caption{Results of 15 tasks (44 planning problems) with 5 repeated tests}
\label{table:results_C:iiwa}
\centering
\begin{threeparttable}
\scalebox{1}{
\begin{tabular}{llcc|cccc|cccc}
\toprule
& \multirow{2}*{Problem} & \multicolumn{2}{c|}{Our method}  & \multicolumn{4}{c|}{numerical optimization} & \multicolumn{2}{c}{probabilistic sampling} \\
& &  \bf{AIP} & \bf{\ffsomp} & {TrajOpt-12} & {TrajOpt-58} & {GPMP2-12} & {CHOMP} & {STOMP} & {RRT-Connect} \\ 
\midrule
\multirow{3}*{Scr(\%)} 
& iiwa\_AB & \bf{100} & \bf{100} & 27.5  & 77.5 & 81.25 & 87.5 & {82.5} & \bf{100}\\
& iiwa\_C & 37.50 & \bf{96.75} & 3.75 &  8.75 & 11.25 & 30 & {25} & 72.5 \\
& aubo\_C & 55.00 & \bf{95} & 3.33 &  6.67 & 11.67 & 26.67 & {25} & 73.33 \\
\midrule
\multirow{3}*{Avt(s)} 
& iiwa\_AB & 1.714 & 1.135 & \bf{0.127} & {0.495} & 1.294 & 2.728 & 3.080 & 8.92 \\
& iiwa\_C & 4.284 & 2.335 & \bf{0.232} & {1.108}  & 2.464 & 6.989 & 7.691 & 13.65 \\
& aubo\_C & 4.812 & 2.163 & \bf{0.245} & {1.119}  & 2.610 & 7.588 & 8.640 & 16.02 \\
\midrule
\multirow{3}*{Sdt(s)} 
& iiwa\_AB & 0.362 & 0.298 & \bf{0.034} & {0.076} & 0.845 & 0.988 & 2.540 & 3.613 \\
& iiwa\_C & 1.048 &  0.273 & \bf{0.055} & 0.101 & 1.312 & 1.290 & 5.233 & 4.562 \\
& aubo\_C & 1.101 &  0.327 & \bf{0.057} & 0.106 & 1.377 & 1.354 & 5.494 & 3.974 \\
\bottomrule
\end{tabular}}
\begin{tablenotes}\scriptsize
     \item[1] Scr(\%), Avt(s) and Sdt(s) denote the success rate, average computation time and standard deviation of computation time, respectively. 
     \item[2] iiwa\_AB, iiwa\_C and aubo\_C denote 16 class A\&B problems on LBR-iiwa, 16 class C problems on LBR-iiwa, and 12 class C problems on AUBO-i5, respectively. 
\end{tablenotes}
\end{threeparttable}
\end{table*}

Figure~\ref{fig:iiwa:C3.2} shows how \ffsomp drags a series of in-stuck arms of Figure~\ref{fig:iiwa:C3} out of the bookshelf by the learning-based collision field to grasp a cup located in the middle layer of the bookshelf safely. Meanwhile, Figure~\ref{fig:AUBO:C6.2} shows how AUBO-i5 avoids the collision of Figure~\ref{fig:AUBO:C6} to transfer the green piece between different cells of the storage shelf safely. 

Table~\ref{table:results_C:iiwa} shows \ffsomp gains the highest success rate compared to the others. The learning-based collision field $\hat{c}(\bm{x})$ and AIP help \ffsomp gain the fourth solving rate after TrajOpt-12, TrajOpt-58, and GPMP2-12. Though TrajOpt-58 compensates for the continuous safety information leakage of TrajOpt-12 in iiwa\_AB, it still cannot escape from the local minima contained by the trust region and gains the second lowest success rate just above TrajOpt-12 in iiwa\_C/aubo\_C. In contrast, \ffsomp successfully descends into an optimum with adaptive primal-dual steps and an appropriate bound of step-size. Thanks to HMC, randomly gaining the Hamiltonian momenta, CHOMP approaches the optimum with the third-highest success rate, whose failures are informed by the deterministic rather than the stochastic gradients. RRT-Connect with a limited time has the highest and the second-highest success rate in iiwa\_AB and iiwa\_C/aubo\_C, respectively. However, a higher rate needs a smaller connection distance, which restricts the RRT growth and computation efficiency.  
As for GPMP2-12 and TrajOpt-58, the LM and trust-region methods help approach the stationary point rapidly. In contrast, the point of iiwa\_C/aubo\_C has significantly lower feasibility than that in iiwa\_A because the initial trajectory of iiwa\_C/aubo\_C gets stuck deeper. 

The comparison between AIP and FFS-OMP in Table~\ref{table:results_C:iiwa} shows how the learning-based method performs a 45\% accelerated descent towards an optimum. 
The 90\% lower success rate of AIP than FFS-OMP shows the limitation of the normal numerical method. It validates that SVM can modify the manifold with less local minima by finding the convex hyperplane. 
 

\section{CONCLUSIONS}

This paper proposes a novel planning method for manipulator motion that utilizes finite Fourier series and optimizes the trajectory numerically informed by the learning-based collision field. 

(i) By adjusting the motion harmonics of each joint, they can represent time-continuous motion more efficiently and effectively. To calculate the Hamiltonian of the manipulator motion harmonics, they sum up both potential energy for collision detection and kinetic energy.

(ii) To reduce shifting between the kinetic and potential parts of the functional objective, we adopt the adaptive gradient method and update exponential moving averages of the gradient and quasi-Hessian. 

(iii) Due to the non-convexity of the collision field, local minima can still be encountered even with an adaptive interior-point method designed to modify harmonics in its finite frequency domain. To address this issue, a support vector machine with a Gaussian kernel is used to learn the highly convex collision field and elevate the success rate by 158\%.




\section*{ACKNOWLEDGMENT}

This work is supported by the Key R\&D Program of Zhejiang Province (2020C01025, 2020C01026), the National Natural Science Foundation of China (52175032), and Robotics Institute of Zhejiang University Grant (K11808).



\bibliographystyle{IEEEtran}
\bibliography{FFS-OMP}

\end{document}